\documentclass[runningheads]{llncs}
\usepackage[T1]{fontenc}
\usepackage{graphicx}
\usepackage{tabularx}
\usepackage{microtype}
\usepackage{subfigure}
\usepackage{booktabs}
\usepackage{hyperref}
\usepackage{csquotes}
\usepackage{adjustbox}
\usepackage{placeins}
\usepackage{multirow}
\usepackage{multicol}
\usepackage{pifont}

\usepackage[capitalize,noabbrev]{cleveref}

\begin{document}
\title{LLMs for LLMs: A Structured Prompting Methodology for Long Legal Documents}
%
%
\author{Strahinja Klem\inst{1}\orcidID{0000-0002-6965-4868} \and
Noura Al Moubayed \inst{1}\orcidID{0000-0001-8942-355X}}
\authorrunning{Klem et al.}
%

\titlerunning{LLMs for LLMs}

\institute{Department of Computer Science, Durham University, Stockton Road, Durham, DH1 3LE, United Kingdom.}

%
\maketitle              
\begin{abstract}
The rise of Large Language Models (LLMs) has had a profoundly transformative effect on a number of fields and domains. However, their uptake in Law has proven more challenging due to the important issues of reliability and transparency. In this study, we present a structured prompting methodology as a viable alternative to the often expensive fine-tuning, with the capability of tacking long legal documents from the CUAD dataset on the task of information retrieval. Each document is first split into chunks via a system of chunking and augmentation, addressing the long document problem. Then, alongside an engineered prompt, the input is fed into QWEN-2 to produce a set of answers for each question. Finally, we tackle the resulting candidate selection problem with the introduction of the Distribution-based Localisation and Inverse Cardinality Weighting heuristics. This approach leverages a general purpose model to promote long term scalability, prompt engineering to increase reliability and the two heuristic strategies to reduce the impact of the black box effect. Whilst our model performs up to 9\% better than the previously presented method, reaching state-of-the-art performance, it also highlights the limiting factor of current automatic evaluation metrics for question answering, serving as a call to action for future research. However, the chief aim of this work is to underscore the potential of structured prompt engineering as a useful, yet under-explored, tool in ensuring accountability and responsibility of AI in the legal domain, and beyond.

\end{abstract}
\section{Introduction}

Large Language models have garnered a great deal of interest beyond the AI community, with their impressive performance across tasks. Going beyond natural language processing, they are now used in visual \cite{carolan2024review,cao2024survey,zhou2024survey} and multimodal \cite{yin2023survey,zhang2024mm} settings as well. In practice, use cases span medicine \cite{thirunavukarasu2023large,muftic2023exploring}, financial technology \cite{zhao2024revolutionizing,wu2023bloomberggpt}, education \cite{yan2024practical,moore2023empowering}, recommender systems \cite{hou2024large,zhao2023recommender}, code generation \cite{liu2024your}, bibliometrics \cite{cano2023unveiling}, and even research itself \cite{birhane2023science}. Yet, despite being a conducive environment, legal advances have not seen the same explosive growth \cite{lai2023large}, in part due to concerns over the unauthorised practice of law, also by instances of misuse. Additionally, Lai et al. point to scalability and credibility as key factors \cite{lai2023large}, whilst Pasquale et al. dismiss the idea in its entirety \cite{pasquale2019rule}. We recognise the necessity for trustworthy tools and methods which augment and facilitate the capabilities of legal practitioners in an accessible way, whilst ensuring their agency and responsibility are always prioritised. \textbf{Thus, we firmly hold that AI models must remain tools, and not become decision makers in their own right.}


We tackle two ubiquitous issues with LLMs which are especially potent within law. Firstly, The long document problem \cite{shukla2022legal}, i.e. when an input document is too large to fit within the context window of a model and cannot be processed entirely. Additionally, even in cases where models are modified to fit longer documents \cite{su2024roformer,peng2023yarn}, they often either underperform outside of their original context, or only perform well near the extremes; known as the lost-in-the-middle problem \cite{liu2024lost}. Secondly, the information retrieval problem \cite{locke2022case}, i.e. the task of retrieving a specific piece of information from a given document or group of documents, such as question answering. Combined, these are amongst the most mundane and repetitive parts of a lawyer's job, which might be better utilised elsewhere. This study addresses both by showing that a correctly prompt engineered LLM, applied in conjunction with other methods, has comparable performance to traditional fine-tuned models.

Our contribution is the introduction of a prompt engineering methodology, whose outputs significantly improve the performance of the generalist model QWEN-2 on the CUAD legal question-answering dataset \cite{hendrycks2021cuad}. We present several techniques that allow a model to tackle documents that exceed their maximum input length. Firstly, we introduce a chunking and augmentation step, where long inputs are converted into smaller chunks, with added redundancy to avoid contextual loss due to separation. By prompting the model at each segment, the search space for question-answering is reduced, increasing reliability and accuracy, but also presenting the task of candidate-selection. Although appearing problematic, this simplifies the task, allowing it to be tackled iteratively from multiple angles. Secondly, we create two heuristics which assess the outputs and isolate the most likely answer. The use of human-understandable methods inherently aids transparency, combating the black-box nature of GPT models. The result addresses all of the previously mentioned problems, and produces a model which is competitive with the state-of-the-art (SOTA).

\section{Related Works}
\label{sec:Related Work}

This section provides a view of the available literature that pertains to the functional components of our work. Namely, the available legal datasets and their limitations, the fine-tuned and generalist GPT models on offer, as well as a look into the various applications and approaches of prompt engineering. We note the seeming disparity between the abundance of data available as compared to models, particularly those in the English language.

\subsection{Datasets}
\label{sec:LegaleseDetails}

From a linguistic perspective, legal data takes a form substantially different from everyday English, often dubbed Legalese \cite{butt2001legalese}. It commonly applies everyday vocabulary outside of the ordinary meaning, makes use of words and phrases from Latin, French and older forms of English\cite{charrow1982characteristics}. Additionally, Legalese often makes deliberate use of either very flexible language with many interpretations or extremely careful and precise phrasings\cite{charrow1982characteristics}. It may be considered semi-structured, being divided into sections and clauses of many types, with the exact length, content and form highly dependent on the legal context. These complexities make tasks involving legal data both challenging and interesting.

There are a wealth of datasets found in the literature. The relevant characteristics considered in this study are the following: \textbf{Jurisdiction.} Within this work, the term is defined as a \enquote{Territory within which a court or government agency may properly exercise its power} \cite{cornellJurisdiction}. Jurisdiction allows us to distinguish between data from countries which share a primary language. Failing to do so carries the risk of misusing or misappropriating statues, cases and terminology between legal systems. \textbf{Intended Purpose/Task.} This includes different forms of information retrieval, judgement prediction or explanation, pre-training and others. Whilst mostly applicable across jurisdictions, certain tasks may only exist or be relevant within certain legal systems.\textbf{Legal System.} The two most typically used systems are common law (case law) and civil law (continental law)\cite{Cohn_1935,law2016common}. Most countries rely on only a single system, however notable exceptions exist (such as China's \enquote{one country, two systems} policy \cite{so2011one})

We showcase these and briefly describe the ones which are curated and associated with a paper, below. An extensive list of non-curated data in other languages and jurisdictions may also be found here \cite{Openlegaldata}.

\begin{table}[ht] 
\label{tab:dataset}
\caption{Most prominently available datasets for law.}
\vskip 0.15in
\begin{center}
\adjustbox{scale=1}{
\begin{tabular}{|l|c|c|c|}
\toprule

Dataset & Jurisdiction & Purpose & System\\
\midrule
CAIL2018   & China&  Judgement Prediction & Civil Law \\
\hline
CAIL2019-SCM   & China&  Case Matching & Civil Law\\
\hline
LeCaRD & China&  Case Retrieval & Civil Law\\
\hline
LexGlue & USA,EU&  Multi-Classification & Case Law\\
\hline
ILDC & India&  Prediction/Explanation & Mixed\\
\hline
BSARD & Belgium&  Article Retrieval & Civil Law\\
\hline
MIRACL  & Many & Retrieval & Many\\
\hline
HAGRID & Many&  Retrieval/Pre-training & Many\\
\hline
MultiLegalPile  & Many & Pre-training & Many\\
\hline
LCR dataset & USA & Contract Review & Case Law\\
\hline

\end{tabular}}
\end{center}
\vskip -0.1in
\end{table}

CAIL2019 \cite{xiao2019cail2019} and LeCaRD \cite{ma2021lecard} are among the largest available datasets from a single jurisdiction, originating from extensive Chinese legal records, and deal with judgement prediction and case retrieval respectively. Meanwhile, LexGLUE \cite{chalkidis2021lexglue} is a multifaceted benchmark dataset, focusing on several multi-label classification tasks, primarily in the English language, from the USA and Europe. English records in the Indian jurisdiction are covered in the ILDC dataset \cite{malik2021ildc}, with a focus on judgement prediction and explanation. Similarly to CUAD, the Lease Contract Review (LCR) dataset \cite{leivaditi2020benchmark} is based in American contract review law, albeit of significantly smaller size (179 documents, compared to the 510 of CUAD).

Aside from these, also notable are the aggregate datasets which combine many different legal records, such as MIRACL \cite{zhang2023miracl} and MultiLegalPile \cite{niklaus2023multilegalpile}, and offshoots like HAGRID \cite{kamalloo2023hagrid}. They contain vast amounts of general legal data that can be utilised for demanding tasks, such as pre-training or fine-tuning of legal models.

\subsection{Models}

In the introduction, we alluded that the scope of available pre-trained LLMs in law is mixed. Whilst there are many diverse implementations available in Chinese \cite{lai2023large,zhang2024citalaw,sun2024lawluo}, considerably fewer models exist for English \cite{colombo2024saullm,chalkidis2020legal}.

Within the former are the following, in chronological order. Firstly, LawGPT \cite{nguyen2023brief}, consisting of the same framework as GPT-3, with the substitution of the RLHF mechanism for domain specific data during training. To our knowledge, this was also the first fine-tuned public legal model to use an LLM as a foundation model. It is followed by ChatLaw \cite{cui2023chatlaw} which expands on this by increasing the size of the model to 13 billion parameters (up to 33 billion in their more recent work), reintroducing RLHF, as well as a KeywordLLM module to address hallucination during information retrieval. The LawyerLlama model \cite{huang2023lawyer} improves on the results of \cite{cui2023chatlaw} whilst also introducing specially designed supervised fine-turning tasks to further bolster performance. According to the authors, the last step in particular allowed their model move from simple legal reasoning to complex chain of thought tasks as well. More recently, CiteLaw \cite{zhang2024citalaw} tackled the problem of reliability and explainability in the domain by introducing a RAG-style \cite{gao2023retrieval} system which can cite articles whilst generating separate responses for laymen and specialists. Meanwhile, LawLuo \cite{sun2024lawluo} extends RAG by applying a Mixture-of-Experts \cite{cai2024survey} approach, taking direct inspiration from Chinese law consultant teams.

Within the realm of English-based models, the authors of LegalBERT \cite{chalkidis2020legal} questioned and explored early approaches to fine-tuning BERT for the legal domain, noting that the contemporary guidance for fine-tuning was ineffective. By contrast, with LegaLMFiT \cite{clavie2021legalmfit}, Clavie et al. showcase the overhead found in transformer-based models, and suggest that LSTMs may be considered for such tasks as well, albeit with reduced effectiveness on complex tasks. However, SaulLM-7B \cite{colombo2024saullm}, based on the Mistral 7B foundation model, incorporates a novel instructional fine-tuning methodology which utilises and expands the LegalBench \cite{guha2022legalbench} benchmark. The authors followed up this work with larger versions of Saul \cite{colombo2024saullm54-141}, but this is, to our knowledge, the only instance of a fine-tuned legal LLM comparable to its Chinese counterparts.

\subsection{Prompt Engineering and Other Methods}\label{related_PE}

Prompt engineering \cite{sahoo2024systematic} is a framework and methodology for structured search, used to craft a prompt which maximises the performance of a given LLM on a specified task. Analogously to software engineering, the employed strategy and output effectiveness depend on the complexity of the task, domain and model knowledge, the type of prompting being applied \cite{wei2022chain} and the specifics of the desired outcome \cite{amatriain2024prompt}. Taxonomies \cite{oppenlaender2023taxonomy,santu2023teler}, best practices \cite{wang2023prompt,white2023prompt}, and frameworks \cite{lo2023clear,sahoo2024systematic} have been created to facilitate a structured and systematic approach which can be applied across different use cases. In terms of real world application, prompt engineering has been widely adopted in industry due to its ease of use, low cost, and lack of training data required compared to fine-tuned models \cite{li2021prefix}. Examples include fields such as academic writing \cite{giray2023prompt}, vision tasks \cite{wang2023review}, job classification \cite{clavie2023large}, business process management \cite{busch2023just} and education \cite{heston2023prompt}. There are also notable examples of legal prompt engineering. In \cite{ribary2023prompt}, the authors utilise various versions of GPT with a rule-based keyword matching algorithm to perform legal document retrieval. However, they opted for a continuous prompting method \cite{li2021prefix}, rather than the discrete ones we have thus far mentioned. These differ in that the former is human uninterruptible input, generated via a separate embedding model, whilst the former incorporates natural language directly. In another study \cite{yu2022legal}, the authors experiment with different prompt types (zero, single and few shot) and focus on Chain of Thought (CoT). They use a GPT-3 model with an extensive Japanese legal dataset \cite{kim2022coliee} for the legal entailment and retrieval problems. Especially notable is that they utilise widely taught legal reasoning strategies, such as IRAC (Issue, Rule, Analysis, Conclusion) \cite{burton2017think} and its modifications as prompting strategies for their model to great effect, reaching SOTA performance on the task. Finally, in \cite{trautmann2022legal}, Trautmann et al. follow a similar methodology to our own, but differ substantially on important points. Firstly, in their study, the dataset was based on Swiss and EU jurisdiction, rather than the American one used in ours (CUAD). Secondly, although long legal documents are used in their work, they are underutilised, as they truncate the document to the length of the model's input context (2048), ignoring the rest. At present, there exist methods such as Rope \cite{su2024roformer} and Yarn \cite{peng2023yarn} which serve to increase context lengths, whilst jointly many models natively support much higher token limits. Our approach avoids these limitations altogether and attempts to instead reduce the size of the input, whilst maintaining the relevant context. Finally, whilst they do make use of prompt engineering, our methodology covers a wider array of approaches and iterates on this by composing them into multifaceted prompts.

\section{Methodology}

\begin{figure*}[htbp]
    \centering
    \includegraphics[scale=0.20]{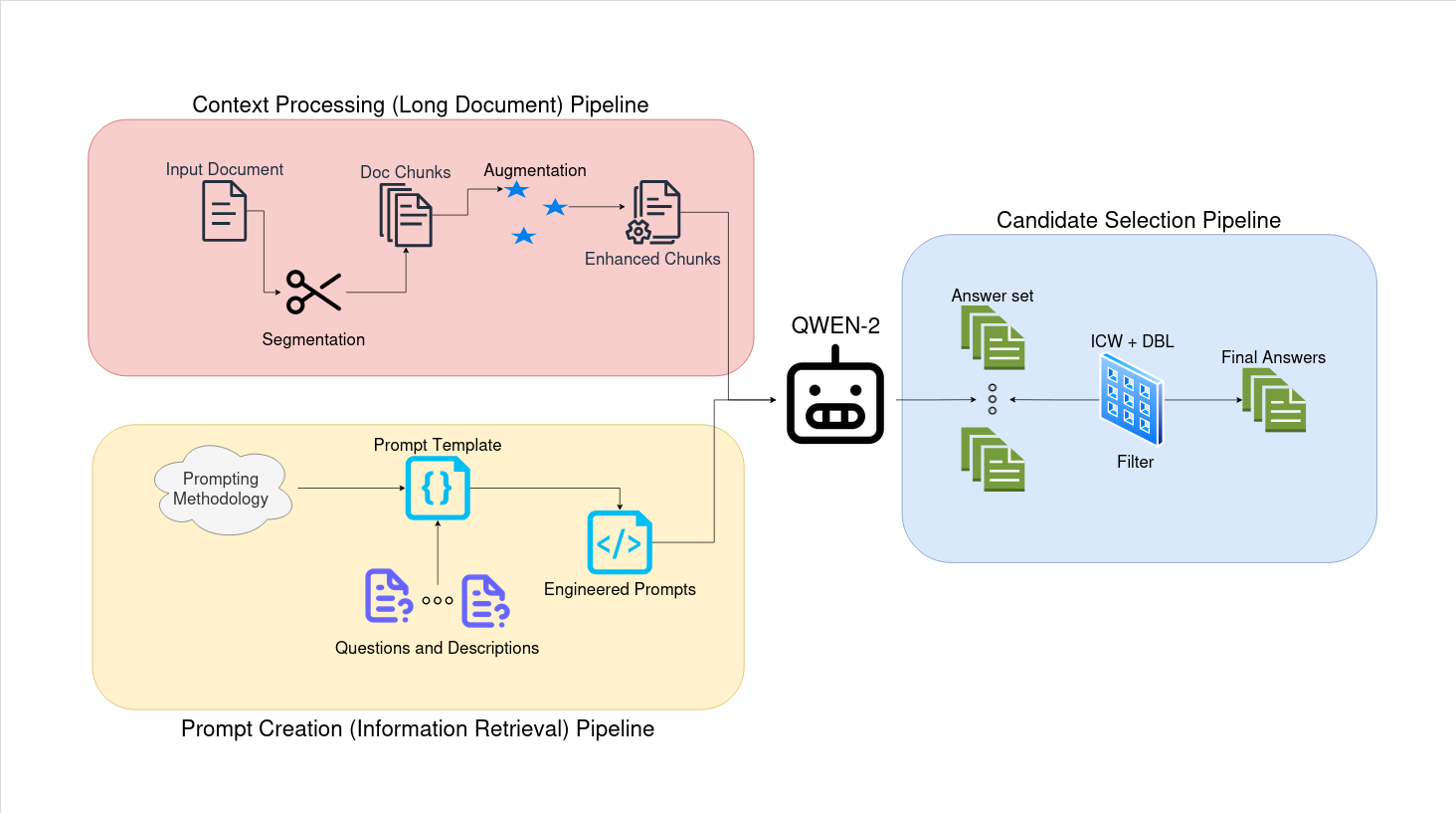}
    \caption{A unified view of our approach with the solutions to long document, information retrieval and candidate selection problems, coloured in red, yellow and blue respectively.}
    \label{fig:unified_project}
\end{figure*}

In this section, we first outline the rationale behind our choice of dataset, model, and evaluation metrics. We then detail the components of our unified approach to long-document question answering and candidate selection, covering chunking and augmentation, prompt engineering, and heuristic refinement. The latter includes Distribution-Based Localisation (DBL) and Inverse Cardinality Weighting (ICW). An overview of the full procedure is provided in \cref{fig:unified_project}.

\subsection{CUAD and QWEN-2}

\begin{figure*}[htbp]
    \centering
    \includegraphics[scale=0.19]{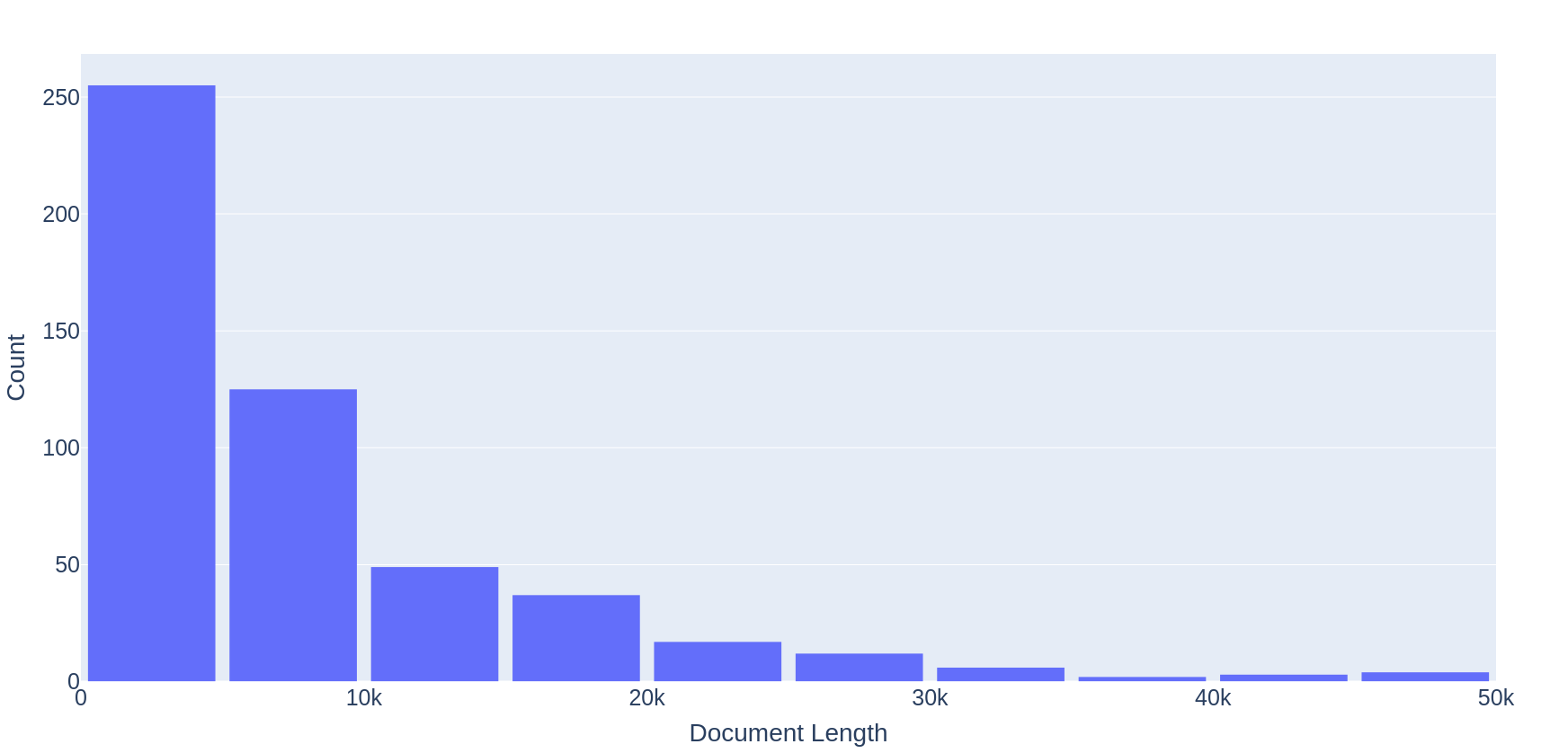}
    \caption{Distribution of document lengths in CUAD.}
    \label{fig:distribution_of_lens}
\end{figure*}

The Contract Understanding Atticus Dataset \cite{hendrycks2021cuad} is an American legal dataset, curated by the non-profit legal organisation "The Atticus Project", with the primary task of information retrieval in the form of contract review. It features 510 documents of various types and lengths (shown in \cref{fig:distribution_of_lens}) , as well as 41 clause-based questions, descriptions and answers for each document, in total representing more than 13000 annotations. CUAD is of notable quality, which is in large part owed to the rigorous training and repeated verification undertaken by the human reviewers, as the authors\cite{hendrycks2021cuad} outline. Although the largest dataset of its kind, datasets in this size range have limited usage for pre-training or fine-tuning, typically requiring hundreds of thousands or even millions of examples, which is an issue that our method avoids. Additionally, data coming from a single jurisdiction removes ambiguity over legal context, working to reduce unnecessary hallucination and improve reliability, which is of critical importance. Due to these qualities, CUAD is a suitable choice for the purposes of our study.

QWEN-2 is an LLM developed and open sourced by Alibaba Cloud, and we ran it locally via the Ollama API. In this study, the 7 billion parameter variant was chosen to serve as a proof of concept, since a reasonable assumption can be made about larger models' performance due to scaling laws \cite{brown2020language}. This decision closely aligns with our stated goal of creating an accessible solution, as our results can be reproduced on a consumer-grade machine rather than a specialised set-up. We use the conversation fine-tuned model, as it allows for greater control over the output than its standard counterpart, and is necessary for all of the prompting strategies that were utilised during the course of the study.

\subsection{Chunking and Augmentation}

The long document problem may be solved by targeting the length of either the context window or the input text. As discussed in \cref{sec:Related Work}, approaches such as Rope and Yarn address the former of the two, usually with a significant trade-off in accuracy and especially outside of the original training context, which is not appropriate for legal question-answering. For a similar reason, although summarisation would fall within the latter group, a naive attempt would suffer from the same lack of accuracy. Ultimately, even for summarisation methodologies  which specifically aim to preserve detail when summarising long text \cite{chang2023booookscore}, the specifics of the legal domain (as in \cref{sec:LegaleseDetails}) present significant and unique challenges. For example, summarising a deliberately vague legal text might result in a specific interpretation being forced. To avoid the above complications, we instead opt for chunking, which does not alter the text of the document. There are multiple approaches that can be taken when segmenting a legal document. One intuitive option would be to split it into sections and paragraphs, thereby conserving the original structure. However, this is not necessarily a straightforward task, and would still present challenges, as legal documents often cross-reference between different sections. Therefore, we opted to split the document in a content-agnostic way, by segmenting into parts of uniform size. This makes the processing of each chunk easier and more consistent, as well as assisting with downstream tasks. The drawback of this method is that it introduces cut-off points which may split key parts of the context, drastically impacting performance. To remedy this issue, we introduce a reduplication step, depicted in \cref{fig:redupfig} below. This addition relinks the context that would have otherwise been lost, significantly reducing the cut-off problem at the cost of additional processing overhead. Although the size of models' context windows is growing sharply, most still have between a range of between 2000 and 8000 tokens, equivalent to approximately 1500 to 6000 words. From \cref{fig:distribution_of_lens}, we can see that while the majority of our documents are of a similar length, a significant fraction is much higher. As well, it is important to remember that the context window is shared with the output, and so should not be in totality taken up by the input. For this reason, chunk size is an important question, and there are several mutually exclusive trade-offs to consider. If the chunks are large, they cover more of the document, retain more of the context, and accommodate longer answers. Additionally, fewer total chunks will need to be queried downstream, saving on computational time and energy. In contrast, smaller chunks create less overhead when reduplicating, but also require more reduplication overall. Importantly, many models have also shown to perform more accurately on smaller inputs, making it more likely to identify an answer if it exists.

\begin{figure}[ht]
\label{fig:redupfig}
\vskip 0.2in
\begin{center}
\centerline{\includegraphics[scale=0.25]{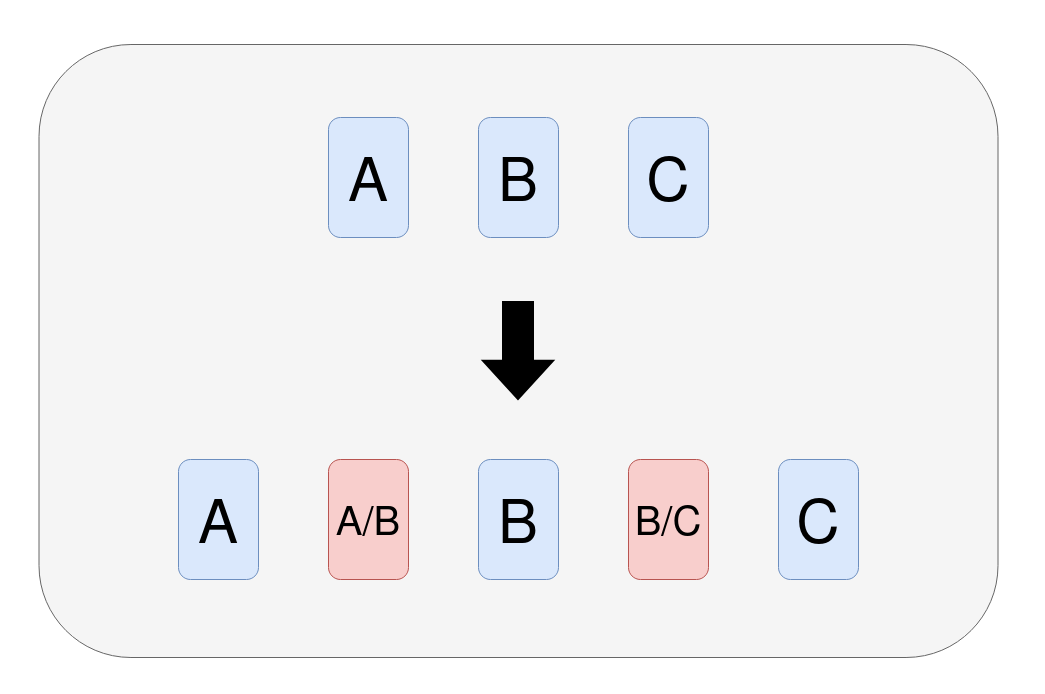}}
\caption{The augmentation (reduplication) step remedies the context splitting problem caused by having cut off points in chunking. This is done by creating a new chunk for every pair of existing sequential chunks such that the new ones contain the latter part of the first chunk and the former part of the second chunk. For example, A/B contains the second half of A, and the first half of B, thereby reconnecting the previously split context.}
\label{fig:redupstep}
\end{center}
\vskip -0.2in
\end{figure}

In this study, we have empirically found that chunk size of 1000 performed the best for our experiments.

\subsection{Prompt Creation and Utilisation}

As we noted in \cref{related_PE}, the process of prompt engineering involves a well structured, systematic approach to prompt creation, testing and optimisation. Our approach is inspired by the training pipeline used in machine learning, with modification where necessary. Below we detail the steps involved in the prompting methodology (highlighted yellow in \cref{fig:unified_project}):

\noindent \textbf{Test and Verification Sets.} As in traditional machine learning, we split the CUAD dataset into two parts, here called the test and validation sets. The former is used to evaluate the average performance of all the candidate prompts, helping to determine which to keep and which to discard. The latter serves to show that the created prompts generalise outside of the test set. When considering a split, a 20/80 stratified split of the dataset would generally be our preferred method, leveraging the efficiency of human optimisation to increase the verification set. However, in the absence of legal expertise, and taking into account the innate difficulty of generative evaluation, we instead adapt to these limitations in the following ways. For the test set, only questions which could be effectively evaluated by a layperson were considered. These questions' categories are document name, parties, agreement date, effective date and expiration date in the dataset. Additionally, to accelerate prompt refinement, only documents of length up to 1000 words (50 total) were selected. As the final prompts were evaluated on chunks rather than full documents, this adjustment still closely follows the intended application. 

\begin{figure*}[htbp]
    \centering
    \includegraphics[scale=0.23]{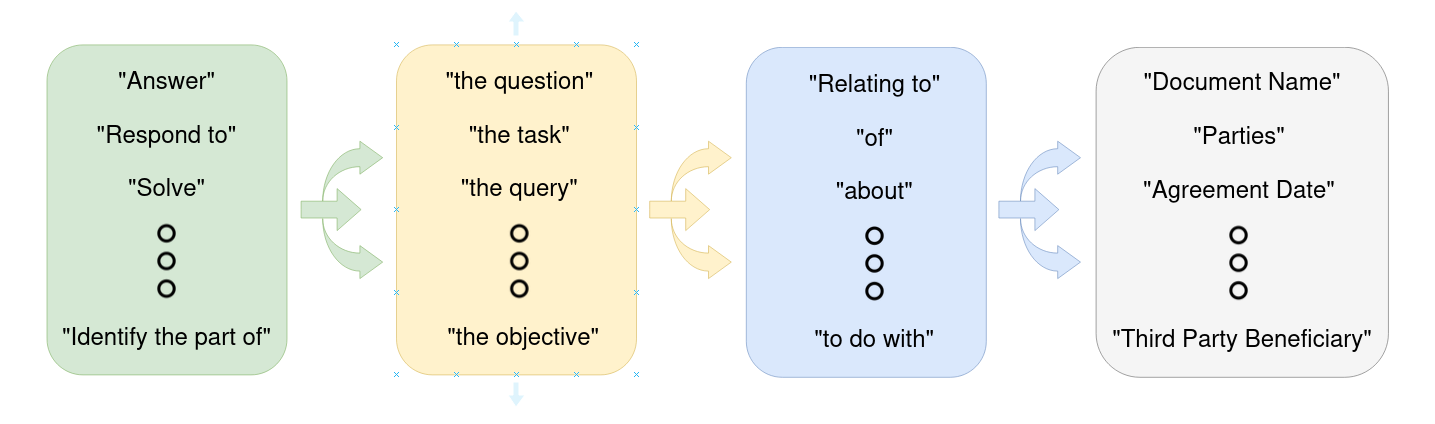}
    \caption{Methodology for creating the base template. Paraphrases are created by taking different combinations from the categories, and are trialled on the test set to evaluate performance. In our experimentation, we found "Identify the part of the question that corresponds to [Q]" to perform the best.}
    \label{fig:BaseTemplatefig}
\end{figure*}

\noindent \textbf{Base Prompt Creation.} In order to generate the initial base prompts, we first decide on a template which can take as input a question from CUAD and direct the model to answer the query. The template can then be modified with synonyms, or rephrased equivalently, to create a pool of paraphrases with the same sentiment. Following this, we evaluate the performance of each prompt on the test set, and select the top performing candidate to become the base prompt template, as in \cref{fig:BaseTemplatefig}.

\noindent \textbf{Technique application.} In line with known methods \cite{white2023prompt,wang2023prompt,ekin2023prompt}, we select all valid combinations of the techniques we believe are most suited to enhance the base prompt template, shown in \cref{tab:prompt-combinations}. An exception was made for the formatting approach, which was applied but not included in the table. Rather than being used to enhance the quality of the output, it is primarily utilised for data cleaning. Thus, the phrase "otherwise respond with 'Does not exist'" is appended to the end of all prompts, both basic and advanced.

Of the other techniques, whilst many are intuitive, there are both inclusions and exclusions which may be unexpected. Of the former, there is some evidence \cite {yin2024should} to show that varying the demeanour of the prompt, i.e. the politeness or rudeness, can influence the quality of the output whilst being relatively nonintrusive for implementation. Additionally, both persona and domain specification techniques were applied, as while both typically aim to achieve the same outcome, the quality of their performance often differs across tasks. As for the latter, we chose to omit few-shot prompting, and related approaches such as active prompting \cite{diao2023active}, as the addition of a second document chunk in the input context would add substantial overhead to an already inference-heavy process. CoT prompting \cite{wei2022chain,zhang2022automatic}, as well as related approaches like self-consistency \cite{wang2022self}, were chiefly left out due to the lack of domain knowledge required to generate legal reasoning, whilst also sharing the previous issue. Following the improvements, the candidates were evaluated a final time via the test set, and the best performing was selected, resulting in the finalised prompt template: 

\enquote{The following text is a excerpt from a larger legal document. If the information is directly present, identify the part of that corresponds to \(Question_i\), otherwise respond only with \textit{"Does not exist"}. In other words, answer the question of \(Description_i\) by quoting it word for word, exactly as it appears in the document, otherwise respond only \textit{"Does not exist".}}

\begin{table}
    {
    \caption{Prompt engineering types and examples}
    \small
	\begin{tabularx}{\linewidth}{|X|X|}
		\hline
		Type & Example \\
		\hline
		Coercive & Base + "or there will be consequences." \\
		\hline
		Kind & "Please " + Base + "Thank you. \\
		\hline
		Intensifier & Base +” as well as possible” \\
		\hline
		Domain & "This is a legal document" + Base \\
		\hline
		Persona & ”Take on the role of a legal expert, and” + Base \\
		\hline
		Rephrasing & Base + ”In other words, " + Question Description \\
		\hline
		Reflection & Base + ”Afterwards, go through it again to improve your response.” \\
		\hline
	\end{tabularx}
    }
\label{tab:prompt-examples}
\end{table}

\begin{table}
    {
    \caption{Valid prompt technique combinations.}
    \small
	\begin{tabularx}{\linewidth}{|X|X|X|X|X|X|X|X|}
		\hline
		Type & Coer. & Kind & Int. & Dom. & Pers. & Reph. & Refl. \\
		\hline
		Coer. & \ding{51} & \ding{55} & \ding{51} & \ding{51} & \ding{51} & \ding{51} & \ding{51} \\
		\hline
		Kind & \ding{55} & \ding{51} & \ding{51} & \ding{51} & \ding{51} & \ding{51} & \ding{51} \\
		\hline
		Int. & \ding{51} & \ding{51} & \ding{51} & \ding{51} & \ding{51} & \ding{51} &\ding{51} \\
		\hline
		Dom.& \ding{51} & \ding{51} & \ding{51} & \ding{51} & \ding{55} & \ding{51} &\ding{51} \\
		\hline
		Pers. & \ding{51} & \ding{51} & \ding{51} & \ding{55} & \ding{51} & \ding{51} &\ding{51} \\
		\hline
		Reph. & \ding{51} & \ding{51} & \ding{51} & \ding{51} & \ding{51} & \ding{51} &\ding{51} \\
		\hline
		Refl.& \ding{51} & \ding{51} & \ding{51} & \ding{51} & \ding{51} & \ding{51} &\ding{51} \\
		\hline
	\end{tabularx}
	}
\label{tab:prompt-combinations}
\end{table}

Prompts for each question can now be generated and paired with every segmented document from the verification set to create the question-answering input for the model.

\subsection{Candidate Selection Heuristics}

After the process of inference is complete, each document will have an answer associated with all of its constituent chunks. This presents the challenge of selecting the correct one, to which end we define the following two heuristics:

\begin{itemize}

\item \textbf{Distribution-Based Localisation.} The overarching idea of this approach is to use labelled data to create a distribution of answer locations over document segments for each question. Doing so, we can approximate a likely location, and more highly weight chunks that contain those segments. It is based on the assumption that legal documents are likely to retain a similar structure as they scale in size, and thus will have clauses (which correlate to answers) in proportionally the same parts of a document. 

The value of 100 for the length of the list L was chosen empirically to balance the granularity of localisation against unnecessary overhead, as well as to create an intuitive interpretation for the partition. We now define the algorithm which generates the final distribution \ $L_f$:

Using the set of documents \( \mathbf{D} \) from the test set, as well as associated answers to questions \( \mathbf{A} \), where each answer \( A \) is a substring of a document \( D \), the following is performed:

\begin{enumerate}
\setlength{\itemsep}{2pt}

\item For every document \( D \in \mathbf{D} \) and every answer \( A \in \mathbf{A} \), define a list \( L \) of real numbers \( l_i \) of fixed length \( |L| = 100 \).

\item Split the document \( D \) into \( |L| \) equal-sized segments \( d_i \), such that \(|d_i| = \frac{|D|}{|L|} \). Simultaneously, segment the answer \( A \) accordingly.

\item Assign values to parts \( l_i \) proportionally based on the presence of \( A \) within \( d_i \),  \(l_i = \frac{|a_i|}{|d_i|}\) where \( a_i \) represents the portion of \( A \) that appears in segment \( d_i \).

\item Construct a set of lists \( \mathbf{L} \) for all documents in \( \mathbf{D} \).

\item Perform element-wise summation across all lists in \( \mathbf{L} \) to obtain \(L_f = \sum_{L \in \mathbf{L}} L\)

\item Normalize \( L_f \) by its maximum value \(L_d = \frac{L_f}{\max(L_f)}\)

\end{enumerate}

This results in a distribution \( L_d \) that approximates the likely location of an answer within a document, leveraging structural regularities in legal texts.

\item \textbf{Inverse Cardinality Weighting.} The aim of the ICW heuristic is to group answers by similarity, and weight them inverse-proportionally to the size of their respective groups. This is done by clustering their GritLM embeddings via the DBSCAN grouping algorithm \cite{ester1996density}, using cosine distance as the metric. This follows from the empirical observation that within the set of outputs generated by the model, the proportion of incorrect to correct answers is very likely to be high, and thus an assumption can be made that correct outputs will appear less often than their counterparts. In execution, this means that all noise points are treated as single element groups which do not penalise the answers, whilst elements of the clusters do.

\end{itemize}

\subsection{Evaluation}

One of the main challenges of this study was to find an appropriate way to compare the original results with our own, as there are subtle, but important, differences between the problem statements of the two, and therefore also their metrics. In the original extractive approach, models are tasked with selecting the entire relevant answer, if it exists, which is supposed to exactly match the form of the ground truth. This makes the task well suited to binary classification metrics (BCMs), such as recall, as outputs can be classified into true or false categories. By contrast, in the generative approach, the task is to produce an answer from the space of all possible responses (rather than only what is written in the document), and check whether it semantically aligns with the ground truth, if it exists. In other words, the output need look nothing like the ground truth as long as the meaning is the same. Furthermore, this means that BCMs no longer apply, as the generative nature of the problem introduces additional possibilities beyond true and false, which cause ambiguities within the categories if applied.  

In fact, evaluation of generative outputs is generally a difficult task, as well as an open problem \cite{van2024field,desmond2024evalullm,chen2024evaluating}. Thus, in this study we employed both automatic metrics as well as human evaluation. The former included the commonly used ROUGE and METEOR, as well as cosine similarity between text embeddings of the ground truth and output. The model used for this was GritLM which has been shown to align closely with human evaluations \cite{muennighoff2024generative}. The threshold for the decision boundary between correct and incorrect was extrapolated by evaluating the average measurements between examples found in the ParaQA question answering paraphrase dataset \cite{kacupaj2021paraqa} as the most relevant dataset we could acquire. The values for ROUGE, METEOR and cosine similarity were 0.60, 0.68 and 0.79 respectively.

\section{Results }

\begin{figure*}[htbp]
    \centering
    \includegraphics[scale=0.22]{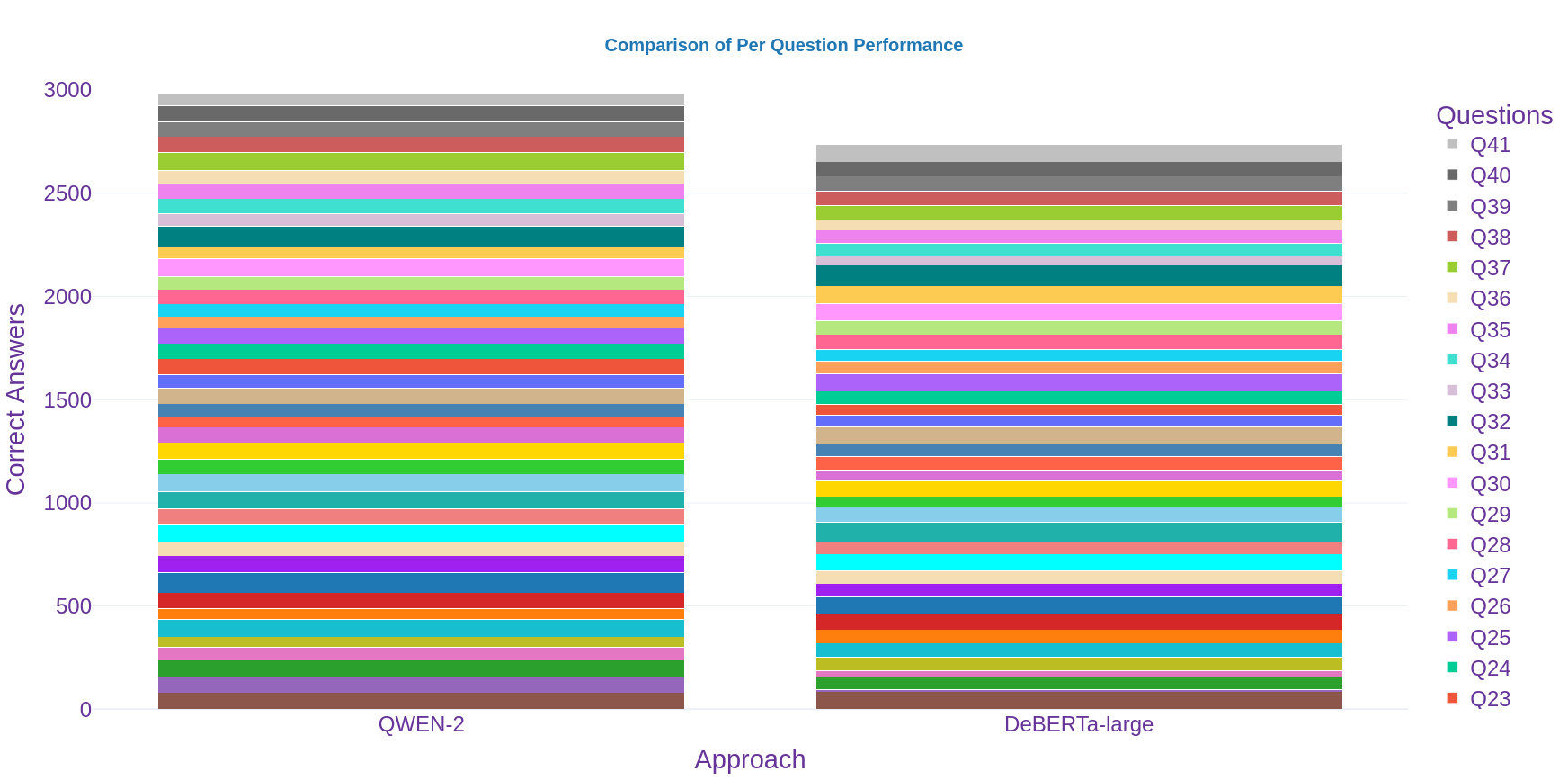}
    \caption{We compare the effectiveness of the two approaches across the 41 questions in the dataset. We find that QWEN-2, with few exceptions, matches or surpasses DeBERTa-large on almost all questions. This represents an average increase in correctness of about 9\% per question, and absolute jump of 250 correct answers total.}
    \label{fig:full_compare}
\end{figure*}

\begin{figure*}[htbp]
    \centering
    \includegraphics[scale=0.22]{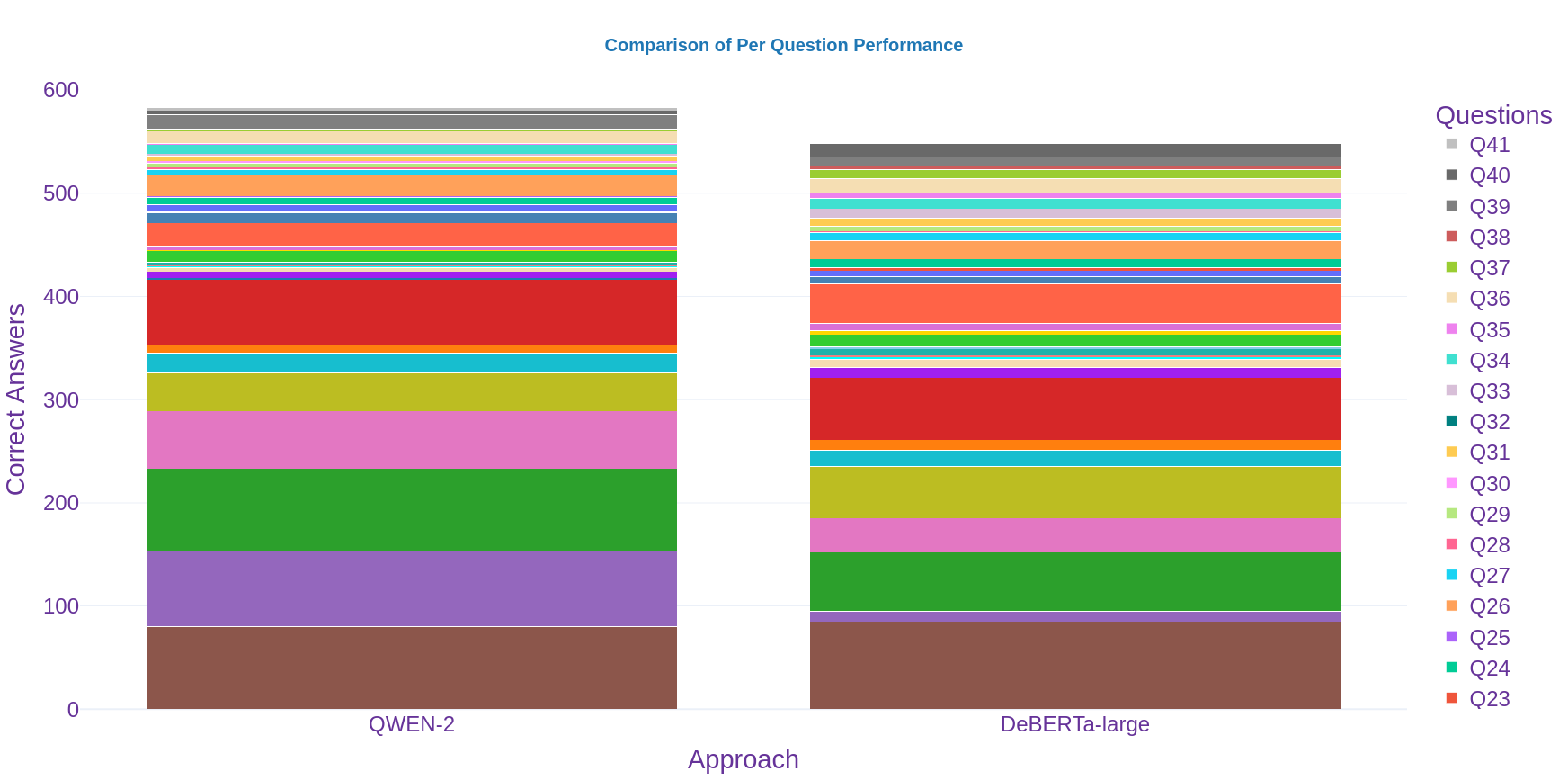}
    \caption{With the same criteria as \cref{fig:full_compare}, we compare QWEN and DeBERTa again, but this time strictly excluding the true negative results. This assessment reflects the primary practical usage of such models by legal experts, whereby the typical concern is whether a clause exists within a document, rather than its absence. We find that QWEN again matches or surpasses DeBERTa on many questions, most notably in Q2. This approach represents an average increase in correctness of about 6\% per question, with an absolute increase of 34 questions.}
    \label{fig:ntn_compare}
\end{figure*}

\begin{table}
\centering
\caption{A full factorial comparison of our approach.}
\small
\begin{tabular}{|l|l|l|}
\hline
Types \textbackslash Measures           & Absolute & Percentage         \\
\hline                   
Basic              & 6775     & 0.324  \\
\hline
Complex            & 7547     & 0.361  \\
\hline
Augmented Basic   & 5608     & 0.268  \\
\hline
Augmented Complex & 10868    & 0.519   \\
\hline
\end{tabular}

\label{tab:FullFactorialExample}
\end{table}

In this study, we endeavoured to show that, by following a structured and optimisation-based prompting methodology on a generalist model, its output can be utilised within a system that produces not only effective, but competitive results on information retrieval from long legal documents. In this section, we showcase and discuss details relating to the following two results. Firstly, the performance of our prompting methodology against that of the DeBERTa model used originally, focusing both the full set of results as well as isolating the true positives. Secondly, a 2x2 factorial design which uses prompting complexity and chunk augmentation as factors, to assess the stand-alone effectiveness of these approaches, as well as their interactions.

During the course of the study, both the automatic metrics (ROUGE, METEOR and cosine similarity) and human evaluation were performed. Comparing the two, we found that whilst these metrics may accurately reflect human judgement when applied to tasks with long output texts, this relationship tends to diverge substantially with text of variable length, such as in our task. Particularly in very short outputs, each difference in syntax or additional text is heavily penalised. Whilst this does not present a problem for DeBERTa's extractive approach, it severely handicaps the generative one. Ultimately, these automatic metrics appear to be too coarse to assess the variable and detailed nature of legal text appropriately. Hence, when comparing QWEN and DeBERTa, we rely on human evaluation to maximise accuracy and reliability, as DeBERTa's 102 document test set is tractable. However, when tackling the results of our different approaches on the whole dataset in \cref{tab:FullFactorialExample}, we instead opt for the automatic metrics.

In \cref{fig:full_compare} and \cref{fig:ntn_compare}, we can see that in both instances, the performance distribution of the two models is similar, with our model performing marginally better overall. However, upon further inspection it becomes apparent that there are multiple failure modes that both models fall into. For QWEN, the most prominent of these are instances where the question is ignored and a summary is provided instead. While DeBERTa cannot generate text, there is an analogous error whereby its selection includes the answer, but far exceeds the required length. Additionally, both models will sometimes give partial answers to questions, although this happens significantly more with DeBERTa. As a generative model, QWEN will occasionally provide correct answers with fallacious or hallucinated reasoning, but amending the prompt to account for this effectively addresses, but does not fully resolve the issue. Likewise, DeBERTa will occasionally be misled by similar but wrong parts of the document, leading to incorrect responses. Our model will also sometimes interpret queries strictly literally, resulting in shallow incorrect responses i.e. answering "Document Name", but is also remedied by modifying the original prompt. Notably, in the question regarding "Parties", there were also instances of false failures in which QWEN outperformed the ground truth. Whilst our model would typically provide all signatories, including alternative aliases, on many occasions the dataset would only offer a partial answer to the question.

In \cref{tab:FullFactorialExample}, we showcase the performance of the prompt engineering and chunk augmentation approaches in their various combinations. The complex prompts outperform their counterparts in all cases, and significantly more so when augmented. This aligns with our expectations, and strongly supports the position that prompt engineering inputs lead to higher performance overall. 

By contrast, duplication does not always correlate with better results, but instead likely amplifies pre-existing tendencies. In the basic augmentation case, where the negative effect is prominent, we suspect an interplay of two primary causes for this. Firstly, the simpler variation of the prompt causes the model to almost exclusively respond in a templated and formulaic way, artificially inducing a higher similarity between correct and incorrect responses. Secondly, if not prompted otherwise, as we do in the complex variant, the model has a very strong bias towards positive outcomes, causing worse performance when the ground truth is negative and and further diluting the pool of outputs. 

These two factors combine to increase the likelihood of DBSCAN grouping the true result with the rest during the ICW heuristic, penalising it during candidate selection. Additionally, this effect is further amplified in the augmented case, whilst also potentially reducing the effectiveness of the DBL heuristic, which amplifies any newly made incorrect outputs in the correct region.

\section{Conclusion }

In this study, we have shown that generalist models, when prompted correctly, can be utilised to combat prominent problems found in the legal domain, even achieving SOTA results in the field. However, whilst the design of our methodology does improve upon the interpretability and robustness of such models, further work is needed to meet the strict standards demanded of AI in the legal domain. Additionally, the challenge of evaluating specialised generative outputs remains an open question, and calls for the development of more specialised automatic metrics.

\bibliographystyle{splncs04}
\bibliography{submission_Mar_2025}

\end{document}